\pdfoutput=1

\documentclass[11pt]{article}

\usepackage[]{ACL2023}

\usepackage{times}
\usepackage{latexsym}

\usepackage[T1]{fontenc}

\usepackage{booktabs}
\usepackage{graphicx}
\usepackage{colortbl}
\usepackage{multirow}
\usepackage{bbding} 
\usepackage{makecell}
\usepackage{bm} 
\usepackage{color} 
\usepackage{xcolor} 
\usepackage{textcomp,booktabs} 
\usepackage{amsmath} 
\usepackage{amsfonts}
\usepackage{amsthm,amsmath,amssymb}
\usepackage{mathrsfs}
\usepackage{subfigure}
\usepackage{mathtools}
\usepackage[utf8]{inputenc}

\usepackage{microtype}

\usepackage{inconsolata}
\usepackage{CJKutf8}
\usepackage{makecell}

\title{Does Correction Remain A Problem For Large Language Models?}

\author{Xiaowu Zhang\textsuperscript{\rm 1}, Xiaotian Zhang\textsuperscript{\rm 2}, Cheng Yang\textsuperscript{\rm 1}, Hang Yan\textsuperscript{\rm 2}, Xipeng Qiu\textsuperscript{\rm 2}\thanks{\;\;Corresponding authors} \\ \textsuperscript{\rm 1}iSchool of North China University of Technology, Fudan University \textsuperscript{\rm 2} \\ \texttt{xiaoxiaowu99@outlook.com, yangcheng\_1212@163.com}
}

\begin{document}
\maketitle
\begin{abstract}
As large language models, such as GPT, continue to advance the capabilities of natural language processing (NLP), the question arises: does the problem of correction still persist? 
This paper investigates the role of correction in the context of large language models by conducting two experiments. 
The first experiment focuses on correction as a standalone task, employing few-shot learning techniques with GPT-like models for error correction. 
The second experiment explores the notion of correction as a preparatory task for other NLP tasks, examining whether large language models can tolerate and perform adequately on texts containing certain levels of noise or errors. 
By addressing these experiments, we aim to shed light on the significance of correction in the era of large language models and its implications for various NLP applications.
\end{abstract}

\section{Introduction}
Large-scale language models are capable of zero-shot learning, addressing natural language processing (NLP) tasks without relying on training data specific to the given task \citep{chowdhery2022palm}. The ability to perform new tasks using prompts is an essential step toward general artificial intelligence. Despite exhibiting remarkable generality and versatility, such as question answering \citep{omar2023chatgpt}, machine translation \citep{jiao2023chatgpt}, and logical reasoning \citep{frieder2023mathematical}, the extent to which large language models (LLM) can correct errors in text and whether they can perform corresponding tasks on data containing erroneous text remains unclear.

\begin{figure}
  \centering
  \includegraphics[totalheight=2in]{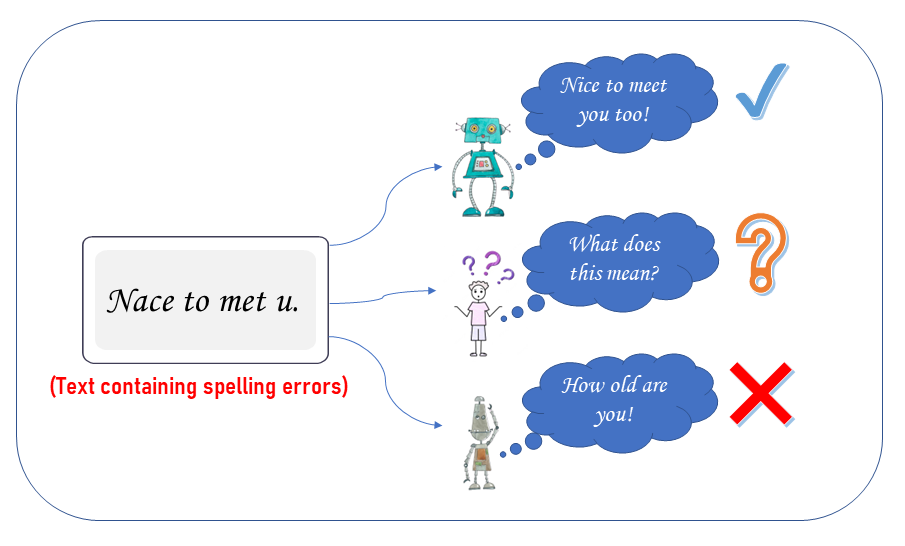}
  \caption{The illustration shows the feedback results of LLM, humans, and other models (such as Bert) when encountering the wrong text. LLM can ignore the wrong text very well. Human beings may be confused when encountering the wrong text, and if the model encounters the wrong text, there is a high probability of error.} 
  \label{fig:graph}
\end{figure}
Text correction has always been an important task aimed at detecting and rectifying text errors, improving language accuracy while reducing the cost of manual verification. \citet{xu2021read} proposed a pre-training model called PLOME, which incorporates spelling error knowledge and uses a GRU network to model the phonetics and strokes of Chinese characters, thereby correcting spelling errors in Chinese text. In English grammar correction, \citet{kaneko2019tmu} utilized BERT to classify ungrammatical and grammatical hypotheses and reranked them based on the classification results. \citet{sun2021instantaneous} introduced Shallow Aggressive Decoding (SAD) to the Transformer decoder and achieved excellent results by combining it with a pre-trained BART model. \citet{rothe2021simple} fine-tuned the multilingual version of the T5 model and achieved state-of-the-art performance in GEC benchmark tests for English, Czech, German, and Russian languages.

This paper evaluates the performance of the text correction task using relevant datasets for Chinese spelling correction (CSC), Chinese text correction (CTC), and English grammar correction (GEC) to determine whether LLM (Language Model) is a qualified text correction model. We assess the corrected results through a combination of human evaluation and self-evaluation using GPT-3.5-Turbo\footnote{\url{https://platform.openai.com/}}, aiming better to determine the model's reliability in text correction tasks. Additionally, we comprehensively evaluate LLM's performance on datasets containing erroneous text by incorporating tasks such as AGIEval\citep{zhong2023agieval}, Gaokao Bench\footnote{\url{https://github.com/OpenLMLab/GAOKAO-Bench}}, machine translation, and sentiment classification. This paper experiments on the GPT-3.5-Turbo model to explore the following two questions.

\textbf{Is LLM an excellent text correction model?} Text correction has always been an unsolved problem. This paper comprehensively evaluates the text correction capability of LLM by examining Chinese spelling correction, Chinese text correction, and English grammar correction. We evaluate the model using three widely adopted evaluation datasets in the current stage. To address the issue of poor data quality in the CSC task dataset, we manually created a dataset containing 1000 evaluation data points for experimentation to assess the performance of LLM in text correction.

\textbf{Can LLM effectively handle tasks involving a certain proportion of erroneous noisy text?} We added a certain proportion of erroneous text noise to the datasets of tasks such as AGIEval [5], Gaokao Bench, machine translation, and sentiment classification to evaluate the extent of fluctuation in LLM's performance. It has been verified that LLM demonstrates tolerance towards data containing noisy information.

After conducting extensive experiments, we have obtained the following results:
\begin{itemize}
\item LLM performs well in text correction tasks and is a highly promising model for text correction. In our evaluation of the CSC task, we found that while LLM has a specific rate of false corrections, it tends to optimize the sentence expression without changing the meaning. Furthermore, the rate of false corrections has been significantly reduced by improving the quality of the dataset and eliminating some erroneous expressions. Meanwhile, for more complex text correction tasks such as CTC and GEC, the increased possibilities in the results lead to a higher rate of false corrections compared to spelling correction. However, manual evaluations also indicate that sentences corrected by LLM are almost free of grammar or spelling errors. All experiments confirm that LLM is an up-and-coming text correction model.
\item LLM performs well in handling data containing erroneous text. The results show that introducing 5\% spelling errors only slightly impacts the experiment's outcomes. As the noise ratio increases, the decline in performance becomes more significant. This demonstrates that LLM can tolerate a certain degree of erroneous text data. However, if noise is introduced by adding or deleting text, which alters the sentence structure, the decline in performance becomes more pronounced.
\end{itemize}
Through a series of experiments, we have demonstrated that LLM is an up-and-coming text correction model and can effectively handle data containing a certain degree of erroneous text. Although LLM has shown some limitations, such as a tendency to optimize sentence structure or expression, its superior zero-shot learning ability and performance on data with erroneous text indicate that LLM is a viable approach toward transitioning from weak artificial intelligence to general artificial intelligence.
\section{Method}
\subsection{Correction}
LLM tends to alter sentence structures to optimize expression when performing text correction tasks. This is partly due to the quality issues in the dataset itself and partly because LLM possesses a solid ability to refine texts. Therefore, this paper adopts a new perspective in evaluating the results. The modification is considered correct if only irrelevant parts have been modified and the sentence's meaning remains unchanged while maintaining grammatical correctness.
\begin{figure}
  \centering
  \resizebox{\linewidth}{!}{
  \includegraphics[totalheight=2in]{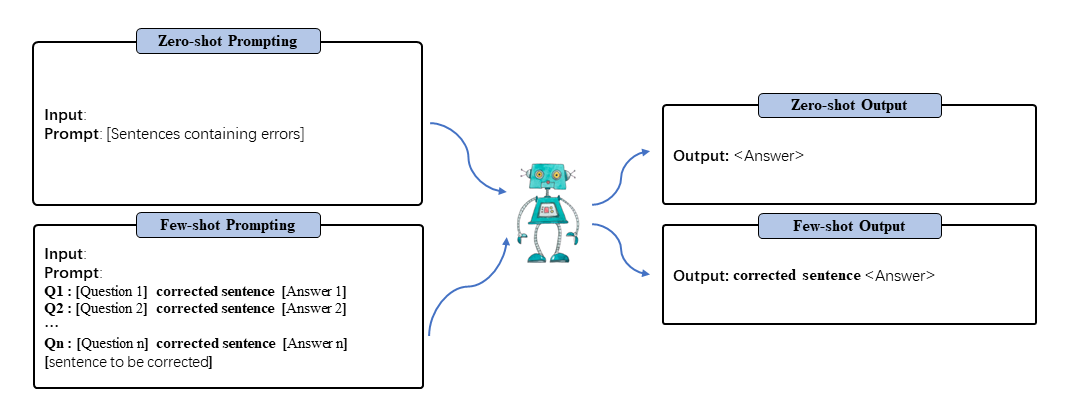}}
  \caption{LLM is guided to perform error correction tasks through zero-shot and few-shots. Zeroshot only uses prompt learning, and few-shot uses in-context learning to provide a case-guided model to complete text error correction tasks.} 
  \label{fig:graph}
\end{figure}
\textbf{Chinese Spelling Correction (CSC):} CSC mainly includes two types of errors: homophonic errors and homographic errors. Due to the large number of sentences with grammatical issues in the dataset, the Language Model (LLM) tends to optimize sentence expression. To mitigate this problem, we introduce additional constraint words in the prompt templates. Although LLM demonstrates remarkable zero-shot capability, incorporating relevant prompt examples can enhance its performance on complex tasks. We provide demonstrations in the prompts to guide the model toward better performance. Following the approach proposed by \citet{min2022rethinking}, we consider both the label space and the distribution of input texts specified by the examples. This study analyzes the error distribution in the validation dataset and selects homophonic and homographic error cases as demonstration examples.

\textbf{Chinese Text Correction(CTC):} Unlike the CSC task, CTC also includes four types of errors: missing errors, redundant errors, mixed sentence structures, semantic repetition, and disorder. CTC focuses on detecting and correcting Chinese spelling errors and grammatical errors. Therefore, the design of prompt templates has been adjusted accordingly. For the few-shot case design, we extracted different error types as demonstration examples based on the distribution of error types, aiming to guide LLM to perform better in the correction task.

\textbf{English Grammar Correction(GEC):} GEC is the task of correcting different types of errors in the text, such as spelling, grammar, word misuse, and punctuation misuse. Similar to the approach used in Chinese text correction, both zero-shot and few-shot approaches were designed for evaluation. The case selection method for GEC is the same as that for Chinese text correction. Examples of different error types, such as replacement errors, omission errors, and insertion errors, were selected based on the dataset as demonstration examples.
\subsection{Task Evaluation with Introducing Noisy Information}

\subsubsection{Noise Sampler}
In order to investigate whether LLM can exhibit remarkable performance on text containing a certain degree of noise or erroneous content, this paper introduces noise information containing erroneous characters into relevant Chinese and English texts. In order to ensure consistency (i.e. keep the same distribution as the original error), we design the noise sampler. It augments the data of related tasks with reasonable noise information.

\textbf{Chinese Text:} Previous research \citep{wang2018hybrid} has shown that phonologically similar and visually similar errors are two main factors contributing to errors in Chinese text, where the number of phonologically similar spelling errors is about two times that of visually similar. Based on the analysis of error distribution in the SIGHAN15 test set, we set the proportions for substituting different types of errors as follows: 60$\%$ for phonologically similar errors, 30$\%$ for visually similar errors, and 10$\%$ for random errors, and applied random substitution to the Chinese text. The character confusion set \citep{wu2013chinese} is a collection of Chinese characters that includes characters with similar pronunciation and similar appearance. Using the character confusion set, we constructed texts with 5$\%$, 10$\%$, and 15$\%$ error proportions.

\textbf{English Text:} We added noise information to the English text using the noise introduction method for constructing the GEC training corpus based on \citet{awasthi2019parallel}. Four types of noise, namely AppendError, VerbError, ReplaceError, and DeleteError, were added according to the distribution of error counts in the available parallel corpus and the probability of each error. First, the number of errors in a sentence was determined through random sampling from {0.4}. Similarly, the selection of each error type was independent and random from {AppendError, VerbError, ReplaceError, DeleteError}. For append, replace, and delete errors, a position was randomly chosen for the error occurrence. In the case of an appended error, the word in that position was dropped. A spurious word from a commonly deleted words dictionary was added to that position for a delete error. For a replacement error, both actions were performed. In the case of a verb error, a verb was randomly chosen from the sentence and replaced with a random verb form of the same word. Commonly deleted words were also obtained from the parallel corpus.
\subsubsection{Task Evaluation}
\textbf{Gaokao Bench:} GAOKAO-Bench is an evaluation framework for assessing large language models' language comprehension and analytical reasoning capabilities, using Chinese college entrance examination (GAOKAO) questions as the dataset. It consists of 1781 multiple-choice questions, 218 cloze questions, and 812 fill-in-the-blank questions. To evaluate the models under the same experimental settings, a certain proportion of noise information is added to the GAOKAO-Bench dataset, and the evaluation is conducted accordingly. In order to ensure consistent evaluation criteria, this paper selects the 1781 objective questions from the dataset and excludes the subjective questions that require subjective evaluation.

\textbf{AGIEval:} proposes a new benchmark, AGIEval, for evaluating the general abilities of foundation models in tackling human-level tasks. AGIEval is specifically designed to assess foundation models in the context of human-centric standardized exams, such as college entrance exams, law school admission tests, math competitions, and lawyer qualification tests. It aims to comprehensively evaluate the model's understanding, knowledge, reasoning, and calculation abilities. In this paper, we selected twenty downstream exam tasks from the open-source AGIEval 1.0 as our evaluation dataset. Similarly, we added corresponding noise information to the dataset. The evaluation was conducted using the same experimental settings.
\section{Experiment}
\textbf{Datasets:} The evaluation data in this paper is selected from the test sets of commonly used datasets in various domains. The Chinese spelling correction task uses the SIGHAN15 \cite{tseng2015introduction} dataset, a commonly used dataset for Chinese spelling correction, consisting of 1100 test samples. 969  samples are used for evaluating the Chinese text correction task\footnote{\url{https://destwang.github.io/CTC2021-explorer/}}, including grammar and spelling errors. In addition, English grammar correction is evaluated using 500 extracted data samples from CoNNL14task \cite{ng2014conll}. For the evaluation of language understanding, logical reasoning, and other abilities of LLM, we utilize the objective question section of the open-source GAOKAO-Bench dataset and the entire dataset of AGIEval 1.0. The sentiment classification task utilizes 1000 samples from the SST-2 English sentiment classification open dataset and 872 samples from the ChnSentiCorp Chinese sentiment classification dataset. We employ the ccmt2019-news dataset consisting of 1011 Chinese-to-English translation samples for machine translation.
\begin{figure}[ht]
  \centering
  \includegraphics[totalheight=2in]{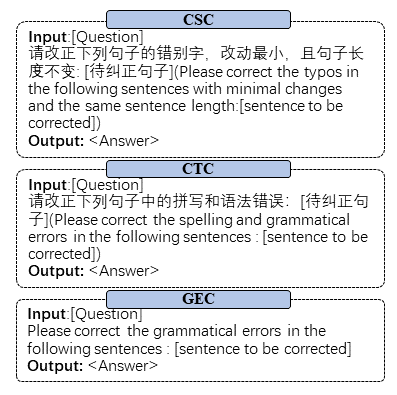}
  \caption{The Mis-correction rate of LLM results as sentence length increases.} 
  \label{fig:graph}
\end{figure}

\textbf{Setting:} For all our LLM experiments, we used GPT3.5-turbo with a temperature setting of 0.02. Reduce the randomness in generating text in language models and produce more conservative text.
Regarding introducing Chinese noise, we employed phonetic-graphemic and random noise methods only for the GAOKAO-bench experiment. For the remaining experiments, we solely used the phonetic-graphemic noise method to introduce Chinese noise. As for English noise, we followed the method described in the previous section to introduce English noise. For the GAOKAO-bench and AGIEval experiments, we utilized their open-source prompts, For the text correction task, we employed prompt templates as shown in Figure 3.
\subsection{Prompt Template for Text Correction Tasks}
\begin{table}[h]
\centering
\resizebox{\linewidth}{!}{
\begin{tabular}{c|c|ccc|ccc}
\hline
\multirow{2}{*}{Dataset} & \multirow{2}{*}{Evaluation method} & \multicolumn{3}{c|}{Zeroshot} & \multicolumn{3}{c}{Fewshot} \\ \cline{3-8} 
                                  &                                             & T         & F         & Acc          & T         & F        & Acc         \\ \hline
\multirow{2}{*}{SIGHAN}         & Human                                       & 994        & 106        & 90.4\%       & 1041       & 59        & 94.6\%      \\ \cline{2-8} 
                                  & GPT                                         & 834        & 266        & 75.8\%       & 905        & 195       & 82.3\%      \\ \hline
\multirow{2}{*}{CSC-Eval}         & Human                                       & 961        & 39         & 96.1\%       & 983        & 17        & 98.3\%      \\ \cline{2-8} 
                                  & GPT                                         & 879        & 121        & 87.9\%       & 893        & 107       & 89.3\%      \\ \hline
\multirow{2}{*}{CTC}              & Human                                       & 853        & 116        & 88.0\%       & 897        & 72        & 92.6\%      \\ \cline{2-8} 
                                  & GPT                                         & 727        & 242        & 75.0\%       & 786        & 183       & 81.1\%      \\ \hline
\multirow{2}{*}{GEC}              & Human                                       & 462        & 38         & 92.4\%       & 473        & 27        & 94.6\%      \\ \cline{2-8} 
                                  & GPT                                         & 447        & 53         & 89.4\%       & 464        & 36        & 92.8\%      \\ \hline
\end{tabular}}
\caption{Results of Text Correction Experiment. T represents the number of correct instances, F represents the number of incorrect instances, and Acc represents the accuracy.Just changed irrelevant parts; as long as the semantics remain unchanged and there are no grammatical errors, \textbf{it is considered acceptable. In this regard, we attempted to use data with added noise in tasks such as AGI EVAL, GAOKAO-bench, and Machine Translation.}, Due to the poor quality of SIGHAN13 and 14 expressions, only 15 were adopted for CSC tasks.}
\end{table}

We employed new evaluation metrics based on different tasks. For GPT-3.5-Turbo, we considered it correct if it corrected a sentence without altering its meaning. We considered it incorrect if it failed to correct the errors or if the corrected sentence underwent significant changes in meaning. We adopted these metrics due to significant issues with the quality of the dataset and the possibility of alternative expressions for each sentence. Therefore, using previous evaluation standards was deemed inappropriate. As shown as Table 1, we found that LLM demonstrated highly competitive performance in the correction task.
\begin{figure}[h]
  \centering
  \includegraphics[totalheight=2in]{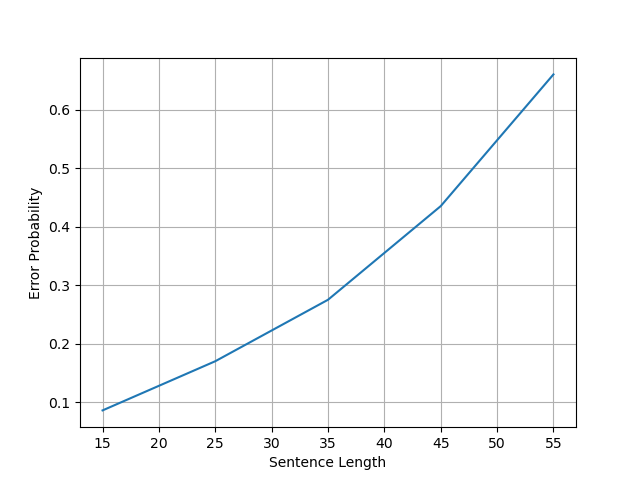}
  \caption{The Mis-correction rate of LLM results as sentence length increases.} 
  \label{fig:graph}
\end{figure}

After conducting manual screening on our dataset, GPT-3.5-Turbo demonstrated excellent performance. This strongly suggests that the high error correction probability is closely related to the dataset's quality. As shown in Figure 4, through the analysis of erroneous sentences, we found that as the sentence length increases, GPT-3.5-Turbo is more likely to make corrections and tends to optimize sentence expressions. This is because longer sentences convey richer meanings, leading to multiple forms of expression that can potentially mislead GPT-3.5-Turbo to correct them into more common expressions.
\begin{figure}[ht]
  \centering
  \resizebox{\linewidth}{!}{
  \includegraphics[totalheight=2in]{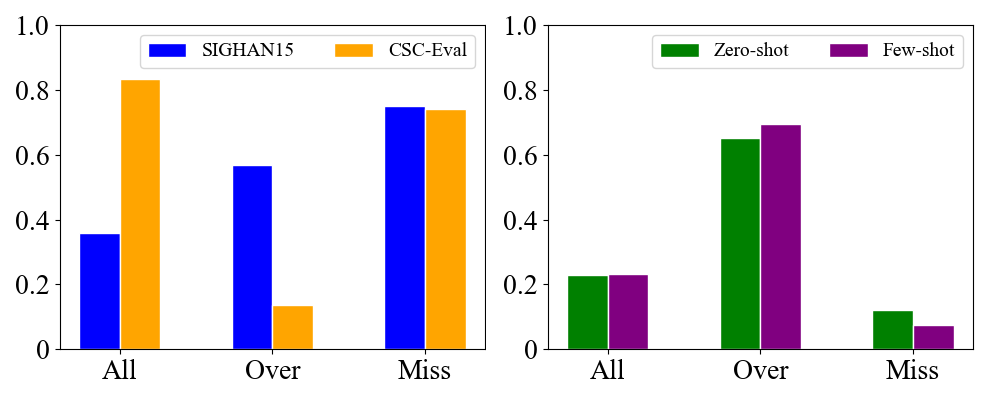}}
  \caption{'All' represents cases where all corrections were correct, 'over' indicates that although errors were corrected, the sentence expression was changed (meaning unchanged), and 'Miss' represents the remaining cases. The left side of the figure compares SIGHAN and CSC-Eval. In contrast, the right side shows the results of the Few-shot and Zero-shot approaches in CTC. The dataset's quality is the main factor affecting error correction, and Few-shot can improve the model's performance.} 
  \label{fig:graph}
\end{figure}

For the Chinese text correction (CTC) task, GPT-3.5-Turbo has also shown promising results. We guide the LLM to correct erroneous sentences by selecting different error cases (in-context learning), and compared to direct zero-shot (only-prompt) error correction, the error correction rate is reduced to 7.4$\%$. Compared to the CSC task, the CTC task is more complex, as it involves requirements such as addition and deletion, and there are more possible forms for correct expressions, which may be a possible reason for its higher over-correction rate.
\subsection{English Grammar Error Correction}
As shown as Table 1, Compared to the CTC task, GPT-3.5-Turbo exhibited relatively better performance on English text. However, upon analyzing the experimental results, we found that the sentences modified by GPT-3.5-Turbo, while altering the expression of the sentences or phrases, maintained correct grammatical structures. Furthermore, we observed that despite the high rate of erroneous corrections in the corrected sentences, the overall meaning of the sentences remained essentially unchanged.
\subsection{Sentiment Classification}
To investigate the impact of text errors on downstream NLP tasks, we introduced noise to the sentiment classification datasets and evaluated the performance of LLM before and after incorporating Chinese and English noise using zero-shot and few-shot approaches. We introduced Chinese and English noise to both the Chinese and English sentiment classification datasets, with a noise ratio of 5\% for the Chinese noise.

\begin{table}[h]
\centering
\resizebox{\linewidth}{!}{
\begin{tabular}{ccccccclllllll}
\hline
\multicolumn{1}{c}{Datasets} & Zeroshot & Zeroshot-noise & Flu             & Fewshot & fewshot-noise & Flu             \\ \hline
SST-2                        & 92.8\%   & 86.0\%   & $\downarrow$5.2\% & 95.8\%  & 91.9\%        & $\downarrow$3.9\% \\
ChnSentiCorp                 & 85.2\%   & 84.8\%         & $\downarrow$0.4\% & 88.9\%  & 88.1\%        & $\downarrow$0.8\% \\ \hline
\end{tabular}}
\caption{Results of Sentiment Classification Noise Experiment."Flu" stands for fluctuation.}
\end{table}

As shown as Table 2, It is evident that after introducing noise, the results on the English sentiment classification dataset SST-2 showed a decrease of 6.8\% and 3.8\% for the zero-shot and few-shot approaches, respectively, compared to the results without noise. On the Chinese sentiment classification dataset ChnSentiCorp, the introduction of Chinese noise had a minimal impact since only character replacements based on phonetic or graphemic similarities were considered, without including insertions or deletions. Therefore, after adding Chinese noise, the zero-shot and few-shot approaches experienced a decrease of 0.4\% and 0.8\%, respectively, compared to the results without noise.
\subsection{Machine Translation}
We further explored the Chinese-English machine translation task to investigate whether GPT-3.5-Turbo performs well without altering sentence structure (i.e., only replacing without involving character additions or deletions). Therefore, using the zero-shot and few-shot approaches, we conducted noise addition experiments using the ccmt2019-news-zh2en dataset with different noise proportions. We evaluated the results using the commonly used machine translation metric BLEU-4 \cite{papineni2002bleu}, averaging the weights for each reference answer. Through the experiments, we found that the variation was minimal when replacing 5\% of spelling errors as noise. In the zero-shot and few-shot approaches, the introduced noise decreased BLEU-4 scores of 0.73 and 0.16, respectively. As the noise proportion increased, at a 10\% noise level, the zero-shot and few-shot results decreased by 2.42 and 2.79, respectively. Under 15\% noise, compared to the no-noise scenario, the zero-shot and few-shot results dropped by 5.29 and 5.35, respectively.
\subsection{AGI Eval with noise}

\begin{table*}[ht]
\centering
\resizebox{\linewidth}{!}{
\begin{tabular}{l|ccl|ccl|ccc|ccc}
\hline
\multicolumn{1}{c|}{\multirow{2}{*}{Task}} & \multicolumn{3}{c|}{Zeroshot}                                                    & \multicolumn{3}{c|}{Zeroshot-CoT}                                                & \multicolumn{3}{c|}{Fewshot}                                    & \multicolumn{3}{c}{Fewshot-CoT}                                 \\ \cline{2-13} 
\multicolumn{1}{c|}{}                      & \multicolumn{1}{c|}{Acc} & \multicolumn{1}{c|}{Acc-N} & \multicolumn{1}{c|}{Flu} & \multicolumn{1}{c|}{Acc} & \multicolumn{1}{c|}{Acc-N} & \multicolumn{1}{c|}{Flu} & \multicolumn{1}{c|}{Acc} & \multicolumn{1}{c|}{Acc-N} & Flu     & \multicolumn{1}{c|}{Acc} & \multicolumn{1}{c|}{Acc-N} & Flu     \\ \hline
aqua-rat                                   & 31.9\%                   & 37.6\%                     & $\uparrow$5.7\%                    & 55.9\%                   & 50.4\%                     & $\downarrow$5.5\%                   & \textbf{\emph{51.5\%}}                   & 48.0\%                     & $\downarrow$3.50\% & 60.6\%                   & 59.2\%                     & $\downarrow$1.40\% \\
math                                       & 26.4\%                   & 19.9\%                     & $\downarrow$6.5\%                   & 31.9\%                   & 27.1\%                     & $\downarrow$4.8\%                   & 14.8\%                   & 9.7\%                      & $\downarrow$-5.10\% & 30.1\%                   & 27.1\%                     & $\downarrow$3.00\% \\
logiqa-en                                  & 35.0\%                   & 37.9\%                     & $\uparrow$2.9\%                    & 39.9\%                   & 36.7\%                     & $\downarrow$3.2\%                   & 43.5\%                   & 42.9\%                     & $\downarrow$0.60\% & 38.9\%                   & 37.6\%                     & $\downarrow$1.30\% \\
logiqa-zh                                  & 41.0\%                   & 40.1\%                     & $\downarrow$0.9\%                   & 38.9\%                   & 37.6\%                     & $\downarrow$1.3\%                   & 46.2\%                   & 46.4\%                     & $\uparrow$0.20\%  & 38.6\%                   & 40.1\%                     & $\uparrow$1.50\%  \\
JEC-QA-KD                                  & 21.1\%                   & 20.4\%                     & $\downarrow$0.7\%                   & 21.2\%                   & 20.8\%                     & $\downarrow$0.4\%                   & 27.6\%                   & 27.7\%                     & $\uparrow$0.10\%  & 23.4\%                   & 23.2\%                     & $\downarrow$0.20\% \\
JEC-QA-CA                                  & 22.0\%                   & 22.2\%                     & $\uparrow$0.2\%                    & 19.6\%                   & 21.1\%                     & $\uparrow$1.5\%                    & 25.1\%                   & 24.8\%                     & $\downarrow$0.30\% & 20.0\%                   & 20.4\%                     & $\uparrow$0.40\%  \\
lsat-ar                                    & 24.4\%                   & 25.2\%                     & $\uparrow$0.8\%                    & 22.6\%                   & 23.0\%                     & $\uparrow$0.4\%                    & 25.7\%                   & 23.5\%                     & $\downarrow$2.20\% & 25.2\%                   & 23.9\%                     & $\downarrow$1.30\% \\
lsat-lr                                    & 52.6\%                   & 50.0\%                     & $\downarrow$2.6\%                   & 52.6\%                   & 52.8\%                     & $\uparrow$0.2\%                    & 59.2\%                   & 53.7\%                     & $\downarrow$5.50\% & 52.2\%                   & 52.1\%                     & $\downarrow$0.10\% \\
lsat-rc                                    & 65.4\%                   & 62.5\%                     & $\downarrow$2.9\%                   & 62.1\%                   & 61.1\%                     & $\downarrow$1.0\%                   & 67.7\%                   & 66.9\%                     & $\downarrow$0.80\% & 57.6\%                   & 56.9\%                     & $\downarrow$0.70\% \\
sat-math                                   & 42.7\%                   & 45.5\%                     & $\uparrow$2.8\%                    & 70.9\%                   & 73.2\%                     & $\uparrow$2.3\%                    & \textbf{\emph{69.0\%}}                  & 66.8\%                     & $\downarrow$2.20\% & 65.0\%                   & 62.7\%                     & $\downarrow$2.30\% \\
sat-en                                     & 81.1\%                   & 78.2\%                     & $\downarrow$2.9\%                   & 77.7\%                   & 79.6\%                     & $\uparrow$1.9\%                    & 81.1\%                   & 82.5\%                     & $\uparrow$1.40\%  & 78.2\%                   & 73.8\%                     & $\downarrow$4.40\% \\
sat-en-no-passage                     & 44.2\%                   & 44.7\%                     & $\uparrow$0.5\%                    & 45.6\%                   & 37.4\%                     & $\downarrow$8.2\%                   & 53.9\%                   & 48.5\%                     & $\downarrow$5.40\% & 51.5\%                   & 49.0\%                     & $\downarrow$2.50\% \\
GK-Cn                                      & 39.0\%                   & 37.5\%                     & $\downarrow$1.5\%                   & 33.7\%                   & 34.2\%                     & $\uparrow$0.5\%                    & 41.5\%                   & 47.2\%                     & $\uparrow$5.70\%  & 37.8\%                   & 35.0\%                     & $\downarrow$2.80\% \\
GK-En                                      & 84.9\%                   & 84.6\%                     & $\downarrow$0.3\%                   & 84.3\%                   & 84.4\%                     & $\uparrow$0.1\%                    & 86.3\%                   & 85.3\%                     & $\downarrow$1.00\% & 84.6\%                   & 81.1\%                     & $\downarrow$3.50\% \\
GK-geography                               & 59.8\%                   & 59.3\%                     & $\downarrow$0.5\%                   & 55.8\%                   & 60.8\%                     & $\uparrow$5.0\%                    & 63.8\%                   & 64.8\%                     & $\uparrow$1.00\%  & 61.8\%                   & 63.3\%                     & $\uparrow$1.50\%  \\
GK-history                                 & 59.7\%                   & 57.9\%                     & $\downarrow$1.8\%                   & 50.2\%                   & 56.2\%                     & $\uparrow$6.0\%                    & 57.6\%                   & 60.1\%                     & $\uparrow$2.50\%  & 58.4\%                   & 61.7\%                     & $\uparrow$3.30\%  \\
GK-biology                                 & 52.9\%                   & 49.1\%                     & $\downarrow$3.8\%                   & 42.4\%                   & 49.1\%                     & $\uparrow$6.7\%                    & 52.4\%                   & 51.9\%                     & $\downarrow$0.50\% & 50.0\%                   & 50.5\%                     & $\uparrow$0.50\%  \\
GK-chemistry                               & 38.7\%                   & 40.2\%                     & $\uparrow$1.5\%                    & 33.8\%                   & 32.9\%                     & $\downarrow$0.9\%                   & 44.0\%                   & 42.1\%                     & $\downarrow$1.90\% & 33.8\%                   & 37.0\%                     & $\uparrow$3.20\%  \\
GK-physics                                 & 33.0\%                   & 29.5\%                     & $\downarrow$3.5\%                   & 29.5\%                   & 33.5\%                     & $\uparrow$4.0\%                    & 33.5\%                   & 32.5\%                     & $\downarrow$1.00\% & 36.5\%                   & 28.5\%                     & $\downarrow$8.00\% \\
GK-Math-QA                                 & 36.5\%                   & 35.3\%                     & $\downarrow$1.2\%                   & 33.3\%                   & 38.2\%                     & $\uparrow$4.9\%                    & 31.3\%                   & 32.5\%                     & $\uparrow$1.20\%  & 31.6\%                   & 33.6\%                     & $\uparrow$2.00\%  \\
GK-Math-Cloze                              & 7.6\%                    & 6.8\%                      & $\downarrow$0.8\%                   & 5.1\%                    & 5.9\%                      & $\uparrow$0.8\%                    & 5.9\%                    & 5.1\%                      & $\downarrow$0.80\% & 8.5\%                    & 9.3\%                      & $\uparrow$0.80\%  \\ \hline
average                                    & 42.9\%                   & 42.85\%                    & $\downarrow$0.05\%                  & 43.2\%                     & 43.62\%                    & $\uparrow$0.42\%                   & 44.4\%                   & 45.85\%                    & $\uparrow$1.45\%  & 45.0\%                   & 44.24\%                    & $\downarrow$0.90\% \\ \hline
\end{tabular}}
\caption{Noise Experiment Results of AGIEval Testing in Zeroshot and Zeroshot-CoT. "Acc" represents accuracy, "Acc-N" represents the results after introducing noise, and "Flu" represents the fluctuation of the results. When we conducted experiments without adding any noise, we found significant discrepancies between some of our results and those provided in the paper. Therefore, we relied on our reproduction results and indicated this portion of the data by bold formatting.}
\end{table*}
As shown as Table 3, The Chinese noise was added at a proportion of 5\%, while the English noise was added based on the probability distribution of error counts. The task types included zero-shot, zero-shot-CoT, few-shot, and few-shot-CoT. We did not modify the original prompts to eliminate the influence of unrelated factors.

We fed the noise-introduced data into the model and obtained output results using the Zeroshot and Zeroshot-CoT approaches. The comparative results are shown in the Table 3. In the Zeroshot setting, we observed a slight decrease in accuracy for 14 tasks, while seven tasks showed a slight increase in accuracy. In the Zeroshot-CoT setting, eight tasks experienced a slight decrease in accuracy, while 13 tasks showed a slight increase in accuracy. However, overall, in the Zeroshot setting, all tasks experienced an average accuracy decrease of 0.05\% when affected by noise. In contrast, in the Zeroshot-CoT setting, all tasks experienced an average accuracy improvement of 0.42\%. The results remain relatively stable.

Similarly, we conducted experiments in the Fewshot and Fewshot-CoT settings, obtaining output results using the original prompts from the dataset. The comparative results are shown in the Table 3. In the Fewshot and Fewshot-CoT settings, we observed a slight decrease in accuracy for 12 and 13 tasks, respectively, while 9 and 8 tasks showed a slight increase in accuracy. When noise was introduced to the data, there were minor fluctuations in accuracy compared to the noise-free data. However, overall, in the Fewshot setting, all tasks experienced an average accuracy improvement of 1.14\% when affected by noise. In contrast, in the Fewshot-CoT setting, there was a decrease of 0.76\%, indicating a relatively stable performance.

\subsection{GAOKAO-bench with noise}
Following the methods proposed in the previous section, we introduced noise into the open-source GAOKAO-Bench dataset for both Chinese and English texts. This was done to test the model's language comprehension ability under conditions of text errors and to quantify the impact of the introduced noise on the model's comprehension ability. We set three levels of noise proportion for Chinese text (5\%, 10\%, 15\%). Additionally, we conducted experiments using the two Chinese noise addition methods introduced earlier to better simulate text errors in real-world scenarios. The task types include zero-shot (only prompt) and few-shots (in-context learning), with all prompts provided by the AGIEval 1.0 dataset without modification.
\begin{table}[ht]
\centering
\begin{tabular}{c|cc|cc}
\hline
N-Noise         & \multicolumn{2}{c|}{P-F-Noise}   & \multicolumn{2}{c}{R-Noise}        \\ \hline
Score-P       & \multicolumn{1}{c|}{Ratio} & Score & \multicolumn{1}{c|}{Ratio} & Score \\ \hline
593                & \multicolumn{1}{c|}{5\%}         & 599   & \multicolumn{1}{c|}{5\%}         & 579   \\ \hline
Score-R & \multicolumn{1}{c|}{10\%}        & 565   & \multicolumn{1}{c|}{10\%}        & 560   \\ \hline
589                & \multicolumn{1}{c|}{15\%}        & 523   & \multicolumn{1}{c|}{15\%}        & 525   \\ \hline
\end{tabular}
\caption{GAOKAO-bench Noise Experiment Results. N-Noise indicates no noise introduced, P-F-Noise indicates PinYin-Font noise introduced. Score represents the total score of objective questions in the experiment. Score-P represents the score mentioned in the paper, and Score-R represents the replicated score.}
\end{table}
We conducted a replication experiment under the same experimental conditions as the original paper. According to the evaluation metrics of the GAOKAO-bench, the experiment used all objective questions, and the results are shown in the Table 4 as the total score of all objective questions. The experimental result was 589 points, which is only 4 points lower than the original paper's score of 593 points, indicating a consistent performance level.

The results of our experiment with phonetic and graphemic noise are shown in the figure. Adding 5\% noise resulted in a score of 599, an improvement of 6 and 10 points compared to the original and replication scores, respectively. However, with a noise proportion of 10\%, the score decreased by 28 points compared to the original and replication scores, and it dropped by 34 points compared to the 5\% noise result. The 15\% noise proportion decreased by 70 points and 66 points compared to the original and replication scores, respectively, and a decrease of 76 points and 42 points compared to the 5\% and 10\% noise results, respectively.

The results of our experiment with random noise are shown in the figure. Adding 5\% noise resulted in a score of 579, a decrease of 14 points and 10 points compared to the original and replication scores, respectively. With a noise proportion of 10\%, the score decreased by 33 points and 29 points compared to the original and replication scores,respectively, and it dropped by 19 points compared to the 5\% noise result. The 15\% noise proportion decreased by 68 points and 64 points compared to the original and replication scores, respectively, and a decrease of 54 points and 35 points compared to the 5\% and 10\% noise results, respectively.
\section{Analysis}
\subsection{Correction }
\textbf{Why is there a discrepancy between the evaluation of GPT-3.5-Turbo in testing and manual assessment? }
\begin{table}[ht]
\centering
\resizebox{\linewidth}{!}{
\begin{tabular}{l|l}
\hline
GPT-3.5-Turbo & \begin{CJK}{UTF8}{gbsn}学生们在老师们的带领下一起探寻徽州文化。\end{CJK}                                                                    \\
Gold Sentence & \begin{CJK}{UTF8}{gbsn}学生们在老师们的带领下一道探寻徽州文化。\end{CJK}                                                                    \\
Translation   & Students explore Huizhou culture together under the guidance of their teachers.         \\ \hline
GPT-3.5-Turbo & \begin{CJK}{UTF8}{gbsn}那个晚上，我睡觉睡得比较安心。\end{CJK}                                                                         \\
Gold Sentence & \begin{CJK}{UTF8}{gbsn}那个晚上，我睡觉睡得比较更安心。\end{CJK}                                                                        \\
Translation   & That night, I slept more peacefully.                                                    \\ \hline
GPT-3.5-Turbo & \begin{CJK}{UTF8}{gbsn}电影结束后，他们就去夜市吃东西，因为他们都很饿。\end{CJK}                                                                 \\
Gold Sentence & \begin{CJK}{UTF8}{gbsn}电影完了，他们就去到夜市吃东西，他们都很饿。\end{CJK}                                                                  \\
Translation    &     After the movie ended, they went to the night market to eat. They were all very                                                                                    \\
         &      hungry. \\ \hline
\end{tabular}}
\caption{Examples of differences between human evaluation and model self-evaluation.}
\end{table}

The disparity in evaluation results is due to GPT-3.5-Turbo's assessment of whether the corrected sentence and the given sentence in the dataset convey the same meaning. Different expressions of the same sentence may mislead GPT-3.5-Turbo in determining whether the corrected sentence matches the intended meaning of the gold sentence in the dataset. For example, in the first sentence, the human evaluation might consider "\begin{CJK}{UTF8}{gbsn}一起探寻\end{CJK}" (explore together) and "\begin{CJK}{UTF8}{gbsn}一道探寻\end{CJK}" (explore along the way) to have the same meaning, but GPT-3.5-Turbo might interpret "\begin{CJK}{UTF8}{gbsn}一起\end{CJK}" (together) and "\begin{CJK}{UTF8}{gbsn}一道\end{CJK}" (along the way) as having different meanings. Similarly, in the third sentence, expressions like "\begin{CJK}{UTF8}{gbsn}电影结束\end{CJK}" (the movie ends) and "\begin{CJK}{UTF8}{gbsn}电影完了\end{CJK}" (the movie is finished), or "\begin{CJK}{UTF8}{gbsn}去夜市\end{CJK}" (go to the night market) and "\begin{CJK}{UTF8}{gbsn}到夜市\end{CJK}" (arrive at the night market) are more colloquial in the dataset. Human evaluators would consider these expressions to have the same meaning, but GPT-3.5-Turbo would perceive them as having different meanings. Furthermore, due to grammatical issues in the original sentences, such as in the second sentence with "\begin{CJK}{UTF8}{gbsn}比较更\end{CJK}" (comparatively more), which is a semantic repetition error, GPT-3.5-Turbo tends to optimize these grammar errors. However, during subsequent evaluation, the model considers the degree of expression different, leading GPT-3.5-Turbo to believe that the two sentences have different meanings.

\noindent \textbf{What are the main reasons for over-correction and under-correction?}
\begin{table}[ht]
\centering
\resizebox{\linewidth}{!}{
\begin{tabular}{l|l}
\hline
GPT-3.5-Turbo & \begin{CJK}{UTF8}{gbsn}有三千名非政府级代表将担任观察员，此外还有二百名记者与会。\end{CJK}                                                                    \\
Gold Sentence & \begin{CJK}{UTF8}{gbsn}有三千名非政府级代表将担任观察员，此外还有二干名记者与会。\end{CJK}                                                                    \\
Translation   &  There will be 3,000 non-governmental representatives serving as observers, in        \\ 
              &  addition to 20,000 journalists attending the event.        \\\hline
GPT-3.5-Turbo & \begin{CJK}{UTF8}{gbsn}他是吃素的人，所以我介绍他台湾的素食。\end{CJK}                                                                         \\
Gold Sentence & \begin{CJK}{UTF8}{gbsn}他是吃素的人，所以我介绍他台湾的青菜。\end{CJK}                                                                        \\
Translation   & He is a vegetarian, so I introduced him to Taiwanese greens.                                                    \\ \hline
GPT-3.5-Turbo & \begin{CJK}{UTF8}{gbsn}作为AI语言模型，我没有具体的时间安排，因此我一直都是有空的。\end{CJK}                                          \\
Gold Sentence & \begin{CJK}{UTF8}{gbsn}你明天有空吗？\end{CJK}                                                         \\
GPT Translation    &     As an AI language model, I do not have specific time schedules, so I am always available.                                                                                    \\
Gold Translation  &  Do you have free time tomorrow? \\ \hline
GPT-3.5-Turbo & \begin{CJK}{UTF8}{gbsn}他们最喜欢的地方是高雄的旗山，那里很漂亮有山有海。\end{CJK}                         \\
Gold Sentence & \begin{CJK}{UTF8}{gbsn}他们最喜欢的地方是高雄的柴山，那里很漂亮有山有海。\end{CJK}               \\ 
Translation    &    Their favorite place is Chai Shan in Kaohsiung, where it is beautiful with both mountains\\ 
        &   and the sea.\\\hline
\end{tabular}}
\caption{Some examples of over-correction and under-correction.}
\end{table}

When conducting manual evaluation analysis, it is easy to observe that GPT-3.5-Turbo tends to optimize the overall structural expression. For example, in the second sentence, it mistakenly corrects "\begin{CJK}{UTF8}{gbsn}青菜\end{CJK}" (vegetable) to "\begin{CJK}{UTF8}{gbsn}素食\end{CJK}" (vegetarian food). It is widespread for certain word expressions to be revised to more common expressions or their hypernyms. Additionally, in the first sentence, GPT-3.5-Turbo believes that out of 4000 representatives from the Ming government, 3000 should be non-government representatives and journalists, so it would change "2000 journalists" to "200" (4000-3000<2000), disregarding the fact that these three categories are not mutually exclusive. Furthermore, in the third sentence, when a sentence to be corrected is a question, GPT-3.5-Turbo is more inclined to answer the question (approximately 40\% of the questions are answered).

In general, using prompts can effectively accomplish the correction task. However, it does not adhere well to requirements such as "do not change the sentence length" or "do not change the meaning of the sentence," and it tends to prioritize optimizing sentence expression. Furthermore, in Chinese spelling correction tasks, GPT-3.5-Turbo is more sensitive to errors with similar pronunciation than to errors with similar character shapes. There is a noticeable increase in over-correction rates for more complex tasks such as CTC and English GEC. The model provides various ways to correct such errors, resulting in answers that may be grammatically correct but differ from the standard answers in the dataset. The results tend to favor common expressions, for example, correcting "\begin{CJK}{UTF8}{gbsn}臆测\end{CJK}"(guess) to the more common "\begin{CJK}{UTF8}{gbsn}猜测\end{CJK}"(guess), even though these two words have different meanings in reality.
\section{Conclusion}
In this paper, we conducted experiments using GPT-3.5-Turbo on relevant text correction datasets to evaluate the performance of LLM in the correction task. Extensive experimental results and analysis have demonstrated the enormous potential of LLM. To address potential issues with the dataset, we manually constructed a Chinese spelling correction dataset consisting of 1000 test samples. Experimental results on this new dataset showed that many cases of over-correction were due to inherent grammar or expression issues in the data. Additionally, in both Chinese CTC and English GEC tasks, although there were instances of over-correction, upon manual inspection of the sentences corrected by GPT-3.5-Turbo, it was found that while the expression and structure of the sentences were altered, the grammar and spelling accuracy was maintained. When adding error noise to the evaluation tasks, we found that LLM handled a certain proportion of spelling errors well. However, significant fluctuations in results were observed when adding or removing text from the original text, which changes the sentence structure. Overall, LLM has tremendous potential in the correction task, and the importance of spelling correction tasks is likely to diminish with the development of LLM.

\bibliography{anthology,custom}
\bibliographystyle{acl_natbib}

\appendix

\end{document}